# Personalization of Itineraries search using Ontology and Rules to Avoid Congestion in Urban Areas

Amir ZIDI*. Amna BOUHANA**. Afef FEKIH***. Mourad ABED*

* LAMIH, University of Valenciennes and Hainaut-Cambrésis, Le Mont Houy, 59313 Valenciennes cedex 9, France
(e-mail: {Mourad.Abed & Amir.zidi}@univ-valenciennes.fr)
** FSEG, University of Sfax, Tunisia
(e-mail: amnabouhana@yahoo.fr)
***University of Louisiana at Lafayette. P.O. Box 43890. Lafayette, LA 70504-3890.
(e-mail: afef.fekih@louisiana.edu )

**Abstract:** There is a relatively small amount of research covering urban freight movements. Most research dealing with the subject of urban mobility focuses on passenger vehicles, not commercial vehicles hauling freight. However, in many ways, urban freight transport contributes to congestion, air pollution, noise , accident and more fuel consumption which raises logistic costs, and hence the price of products. The main focus of this paper is to propose a new solution for congestion in order to improve the distribution process of goods in urban areas and optimize transportation cost, time of delivery, fuel consumption, and environmental impact, while guaranteeing the safety of goods and passengers. A novel technique for personalization in itinerary search based on city logistics ontology and rules is proposed to overcome this problem. The integration of personalization plays a key role in capturing or inferring the needs of each stakeholder (user), and then satisfying these needs in a given context. The proposed approach is implemented to an itinerary search problem for freight transportation in urban areas to demonstrate its ability in facilitating intelligent decision support by retrieving the best itinerary that satisfies the most users' preferences (stakeholders).

## 1. INTRODUCTION

Urban Freight transport has often been neglected in urban traffic management. Though it is very important for the business and life of a city, commercial traffic has not received much attention in the transport planning process.

The last decades have witnessed tremendous efforts aimed at creating and developing a new generation of intelligent freight transportation systems with the objective of controlling and optimizing congestion, fuel consumption, and environment problems and increasing the safety of passenger and goods in urban areas.

The question of efficiency for urban freight transport systems is considered one of the most complicated matters because it involves many actors such as shippers, carriers, receivers, public authorities and citizens. Further, it requires a significantly large and robust information system to manage and insure the operation of freight transportation in order to reduce the complexity of urban freight transport and improve its performance. Thus, a better organization of urban freight traffic can reduce urban congestion, atmospheric pollution and noise while improving security issues for people and goods. One way to overcome some of the problems of urban areas is to propose some solutions to the congestion problem.

Solving this problem requires the availability of reliable and relevant information system about different transportation means used to insure the transportation of goods in urban areas. However, the existence of several actors in city logisics and several operators managing these various transportation means as well as the diversity of information sources and user profiles make the task of designing a homogeneous information system a challenging one. The problem is no longer the availability of information but rather the ability to select the relevant information that meets the precise needs and interests of various stakeholders (residents, retailers-authorities, supplier, carrier) in order to select the best itinerary that minimizes the combination of sets of criteria that satisfy the user needs and interests and overcome the problem of congestion thus improving the operation of distribution network in urban areas.

Hence, personalization in itinerary search has recently become a major goal for intelligent freight transport systems. As a result, several approaches for personalized information systems in the field of transportation, especially for passengers, were recently developed to provide users with information that satisfies their preferences (Letchner *et al*., 2006; Akasaka and Onisawa, 2008, Nadi *et al*., 2011; Bouhana *et al*., 2010; Bouhana *et al*., 2013). Note that all the above mentioned approaches used classical methods of personalization such as content-based or collaborative-based methods.

In recent years, new paradigms for personalization of applications have emerged such as the Ontology concept (Shi & Setchi., 2013) (Olivera *et al*., 2013) (Riano *et al*., 2012). Ontology allows for the sharing of a large group of information sources or knowledge bases, knowledge organization, and interoperability among complex systems. Ontologies are a level of description of the knowledge of a system. The use of ontology offers many advantages such as determining the appropriate criteria for the user and context modeling. In decision problems, the ontology-based approach is employed to incarnate the role of experts or experienced practices in personalized evaluation processes.

In the literature, only few studies have applied ontology-based approaches to the field of transportation. In Niarki *et al*., 2009, the authors developed a new approach for personalized route planning search in public transport based on the combination of the AHP methods and the Ontology. Olivera *et al*., 2013 used ontology to generate personalized user interfaces for transportation interactive. However, all the above mentioned studies have only focused on the personalization of information for passengers in public transport and neglected freight transportation in inter-city or urban areas.

In order to overcome those limitations, we propose in this work to provide a new ontology-based personalization approach for freight transportation in urban areas which satisfies the preferences of all city logistics' actors. The main objective of this research is to design new itinerary search system in order to improve the distribution process of goods in urban areas and also to reduce their consequences such as congestion, fuel consumption, and environmental impact.

In this work, we present a new technique of personalization to mitigate the problems of congestion and reduce its negative consequences. The main contributions of this paper are as follows:

- The design of a new method of personalized itinerary search for freight transportation in urban areas based on ontology and rules in order to alleviate the problem of congestion and its consequences.

- City logistics ontology is used as the representation formalism to model both content sources and user profiles.

- The personalization tasks are solely carried out based on the inference power of the rules.

The rest of this paper is organized as follows. In section 2 we present our new approach of personalization in city logistics based on ontology and rules. The results of the implementation of the proposed approach to an urban transport problem are illustrated in Section 3. Some concluding remarks and proposed future work are discussed in section 4.

## 2. PROPOSED APPROACH

The proposed approach is based on the ontology-based city logistics paradigm GenCLON proposed by Anand *et al*., 2012.

### 2.1. Scope of GenCLON

City logistics is the study of the dynamic management and operations of urban freight transport and distribution systems. The aim is to ensure optimum productivity, reliability and customer service whilst reducing environmental impacts, air pollution emissions, energy consumption and traffic congestion. City logistics is classified as a discipline which can cope with sustainable problems encountered in urban logistics freight transportation. One of its key characteristics is its ability to take into consideration the heterogeneity of stakeholders in the decision process (heterogeneity of information and interest exchanged between different stakeholders (Residents-Retailers-Authorities- Supplier-Carrier). In, (Anand *et al*., 2012), the authors proposed an ontology based city logistic tool (GenCLOn). The aim of this ontology is to formalize the domain knowledge of the city logistics in order to facilitate communication and share different knowledge with several terminologies and types of decisions made by the different actors in the city logistics such as residents, retailers, authorities, suppliers, carriers.

From a semantic point of view Anand *et al*., 2012 used the concept of ontology to develop a sort of glossary or a common language in order to facilitate communication and coordination among the different actors involved in City logistics. The objective of this ontology is to capture the maximum of knowledge about the city logistics domain.

In the City logistics the major objectives or interest are economic; environmental and social which can ultimately influence the urban movement pattern (see Fig.1)

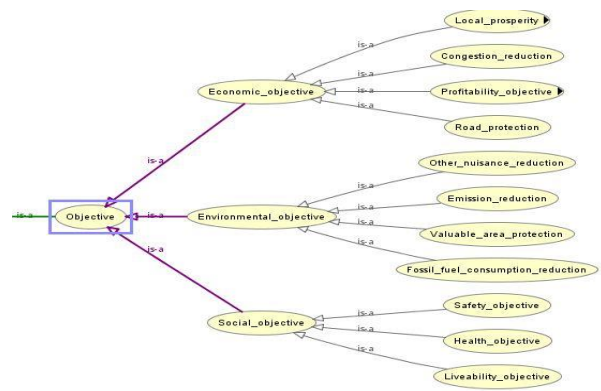

Fig. 1. Taxonomy objectives in GenCLOn (Anand *et al*., 2012)

From the above figure, we can conclude that the competitively and the growth of economy in urban area depend on congestion reduction, nuisance reduction, consumption of fuel reduction.

In order to alleviate the problems of congestion and its consequences in urban areas, we propose here a new solution based on the concept of personalization using the city logistics ontology (Anand *et al*., 2012) and rules in order to optimize a set of objectives in urban freight transportation.

## 2.2 Rules Based Personalization

In general, ontologies concentrate on classification methods, by defining 'classes', 'subclasses' and associating individual resources to such classes, and characterizing the relationships among classes and their instances see Figure.1. Rules, on the other hand, focus on defining a general mechanism aimed at discovering and generating new relationships based on the existing ones.

In our paper, we use rules in order to obtain an efficient application in the field of personalization of itineraries in order to improve the quality of urban freight transport and avoid congestion. In our approach, we define a set of rules using SWRL among GenCLOn (Anand *et al*., 2012). The objective of these rules is to satisfy the user's preferences (user's profiles) in a given context.

, Both private and public users or stakeholders are involved in city logistics with different preferences (Fig.2). For private stakeholders, such as shippers, carriers and receivers, the main objective is to reduce transportation cost and minimize the distance travelled for a given transport mission. For public stakeholders (public authority) and residents, the main objectives are getting better accessibility inside the city with less fuel consumption as well as reduction of the environmental impacts such as pollution, nuisances, etc…

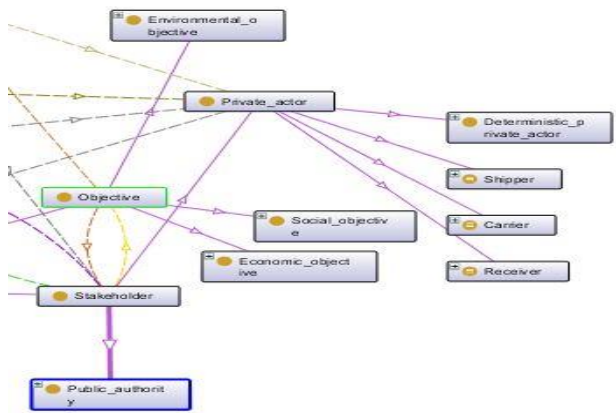

Fig.2. Taxonomy of stakeholders and their objectives

Therefore, in order to satisfy the preferences of all the stakeholders, while reducing the negative impacts of freight transportation, personalization of information has a key role in improving the process of itinerary search Using, GenCLOn ontology as a backbone for our new approach for personalization, we develop an ontology that perfectly represents the city logistics domain as well as all the knowledge about the content sources and user profiles. That is, the combination between ontologies and rules can lead to more efficient personalization. A set of DL-safe rules was written, using SWRL. These rules match user's preferences which mainly represent stakeholders' objectives or resources in a given itinerary. The rule is a conjunction of concepts and axioms that constitute knowledge. In Table.1, we cite some axioms which basically illustrate relations between concepts.

**Table.1 Axioms in GenCLOn (Anand et al.2012)**

| Axioms |
| --- |
| Carrier has_activity some Loading |
| Carrier has_ressource some Driver |
| Carrier has_ressource some Road_freight_vehicle |
| Receiver has_objective some Other_nuisance_reduction |
| Shipper has_objective someTransport_cost_reduction |

Based on the organization of stakeholders, several kinds of itinerary patterns can be offered to the user for a better planning. We organized them in taxonomy of itinerary patterns (Fig.3). In this way, an itinerary pattern is classified as a shipper pattern, a receiver pattern or a shipper-receiver pattern.

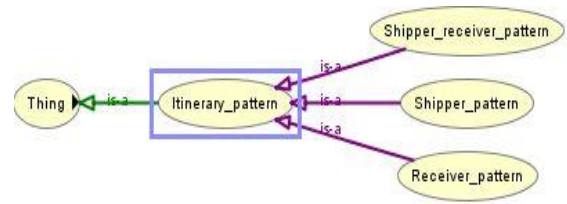

Fig.3. Taxonomy of itinerary pattern

Initially, three SWRL rules were written to formalize the taxonomy of itinerary pattern based on stakeholders' objectives to check whether user preferences were satisfied or not.

An example is to express the fact that "if we have an itinerary in which shipper has an objective, then the itinerary pattern (*Iitinerary_pattern(i?)*) is classified as a shipper pattern *(Shipper_pattern(i?)*. the SWRL description for this rule is depicted in fig.4 in which, the shipper has a transport cost reduction as an objective. The rest of the rule describes the resources used by the carrier.

| SWRL Rule #1 |
|---|
| Transport_cost_reduction(?t), Loading( ?l), Carrier(c?), Itinerary_pattern(?i), Receiver( ?e), Shipper(?s), Driver(?d), Road_freight_vehicle( ?r), has_ressource( ?c, ?d), has_ressource( ?c, ?r), has_objective(?s, ?t)->Shipper_pattern(?i) |

Fig.4. Rule for shipper preferences

In the second rule (Fig.5), the itinerary pattern (*Iitinerary_pattern(i?)*) is classified as a receiver pattern *(receiver_pattern(i?)* since the receiver has other nuisance reduction as an objective.

| SWRL Rule #2 |
|---|
| Other_nuisance_reduction(?o), Loading(?l), Carrier(c?), Itinerary_pattern(?i), Receiver( ?e), Shipper(?s), Driver(?d), Road_freight_vehicle(?r), has_ressource( ?c, ?d), has_ressource( ?c, ?r), has_objective( ?e, ?o)->Receiver_pattern(?i) |

Fig.5. Rule for receiver preferences

For the case where the shipper and the receiver have objectives in the same itinerary, we wrote a third rule (Fig.6).

| SWRL Rule #3 |
|---|
| Transport_cost_reduction(?t), Other_nuisance_reduction(?o), Loading(?l), Carrier(c?), Itinerary_pattern(?i), Receiver(?e), Shipper(?s), Driver(?d), Road_freight_vehicle(?r), has_ressource( ?c, ?d), has_ressource( ?c, ?r), has_objective( ?s, ?t), has_objective( ?e, ?o)->Shipper_receiver_pattern(?i) |

Fig.6. Rule for shipper and receiver preferences

Consequently, a SWRL rule was defined in order to associate the user with all the itineraries that meet exactly his preferences.

## 3. APPLICATION

As mentioned previously, the domain in which the approach was evaluated was urban Freight transport. We use **GenCLOn** core ontology which contains x concepts. In order to verify the approach and illustrate its application in a real world itinerary search, a real data collection is used to build the first knowledge base. It was retrieved from Paris Open Data.

The interface and the functionality of the application are implemented, as usual for web applications, in JEE (Jsp and Servlet). For the ontology extension and development, we use OWL (Ontology Web Language) in order to ensure the maximum possible expressiveness and the efficient support of reasoning. The definition of rules is implemented in SWRL. Protégé 4 has been selected as the appropriate ontology environment to implement our ontology and its associated rules. We make use of Pellet, an open-source Java based OWL reasoner, which guarantees complete OWL reasoning. The special feature of Pellet is its support for reasoning with DL-safe rules which makes it particularly suitable for our needs.

In the proposed system, we adopt keyword-based interface as it provides a comfortable and relaxed way to query about the end user, in which the user sets the addresses of the shipper and the receiver (Fig 4). The system provides also the user with a configuration to set his preferences, in which he can select the desired criteria to be included in personalized itinerary search.

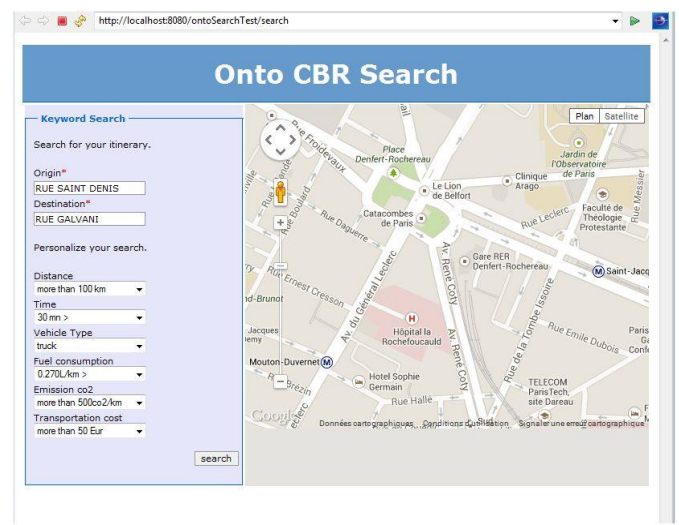

Fig.7 .User interface

In (Fig. 7) we have the query: "go from Rue Saint Denis (shipper) to Rue Galvani (receiver)" with search criteria: Time less than 30 minutes, vehicle type is truck, fuel consumption less than 0.270, emission CO2 less than 500 co2/km and transportation cost less than 50 euro".

Fig.8. Query results

Results are shown in (Fig. 8), we have a set of personalized results which represent real world itineraries. In this case, the stakeholders like shipper (origin) and receiver (destination) has respectively an economic (Transportation cost) and environmental (Gaz emission) objectives and the type of vehicle chosen is a truck. Then, concepts with labels {"Carrier", "Shipper", "Receiver","Road_freight_vehicle", "Transport_cost_reduction", "Emission_reduction"} and properties with labels {"connect", "origin", "destination", "has_objective", "has_resource"} are selected from GenCLOn. Finally, itineraries are calculated based on Rule#3 which contains the selected concepts and properties.

## 4. CONCLUSION

In this paper, we proposed a new solution for improving the process of freight transportation in urban areas, while decreasing the problem of congestion and its consequences. Our approach integrates the concept of personalization in the intelligent freight transportation as a technique to alleviate the problem of congestion in urban areas by providing the users with a personalized itinerary that satisfies exactly their needs and preferences. The proposed approach is based on ontology and SWRL rules. The ontology is used to formalise the representation of various sources of information and to capture knowledge about user preferences in city logistic while the inference power of the rules insures the personalization process.

In our future research, we plan to further optimize the itinerary search by including information about traffic lights and road congestion in real time. This will lead to a dynamic and personalized itinerary search engine for urban freight transport and distribution systems.